# Natural Language Processing Chains Inside a Cross-lingual Event-Centric Knowledge Pipeline for European Union Under-resourced Languages

**Diego Alves, Gaurish Thakkar, Marko Tadić**

University of Zagreb, Faculty of Humanities and Social Sciences
Ivana Lučića 3, 10000 Zagreb, Croatia
dfvalio@ffzg.hr, gthakkar@m.ffzg.hr, marko.tadic@ffzg.hr

**Abstract**

This article presents the strategy for developing a platform containing Language Processing Chains for European Union languages, consisting of Tokenization to Parsing, also including Named Entity recognition and with addition of Sentiment Analysis. These chains are part of the first step of an event-centric knowledge processing pipeline whose aim is to process multilingual media information about major events that can cause an impact in Europe and the rest of the world. Due to the differences in terms of availability of language resources for each language, we have built this strategy in three steps, starting with processing chains for the well-resourced languages and finishing with the development of new modules for the under-resourced ones. In order to classify all European Union official languages in terms of resources, we have analysed the size of annotated corpora as well as the existence of pre-trained models in mainstream Language Processing tools, and we have combined this information with the proposed classification published at META-NET whitepaper series.

**Keywords:** language processing chains, under-resourced languages, European languages resources.

## 1. Introduction

It is indisputable that major events such as Brexit and the recent migration crisis affect countries inside the European Union (EU) in several different ways. Due to the impact of major events inside local communities, with different languages, an enormous amount of event-centric multilingual information is available from different media sources. This diversity reflects community-specific aspects, opinions, sentiments, and bias (Annex 1 to the Grant Agreement of Cleopatra – 812997).

This multicultural data is potentially useful for a great variety of stakeholders, including digital humanities researchers, memory institutions, media monitoring companies, and journalists. However, in order to provide all the presented information in a valuable way, it must undergo first through a sequence of automatic processing: effective interlinking, verification, and analytics. The aim of CLEOPATRA[1] MSC Innovative Training Network (ITN) is to address these needs by bringing the cross-lingual event-centric information analytics technology to a higher level.

To achieve its objective, CLEOPATRA initiative focuses on three main dimensions:

- Alignment, validation, and contextualization of event-centric multilingual information across heterogeneous sources for all twenty-four EU official languages.
- Development of new interactive user access models to cross-lingual information to optimize the way to interact with the diverse data at different levels.
- Development of models that describe cross-cultural information propagation in a data-driven, application-centric manner.

The European Union (EU) has around 513 million inhabitants (Eurostat[2], 2020) and twenty-four official languages: Bulgarian, Croatian, Czech, Danish, Dutch, English, Estonian, Finnish, French, German, Greek, Hungarian, Irish, Italian, Latvian, Lithuanian, Maltese, Polish, Portuguese, Romanian, Slovak, Slovene, Spanish and Swedish (European Union[3], 2020). One of the main challenges of CLEOPATRA ITN is the discrepancy of available data and resources between EU languages. This difference has an impact on the automatic extraction and processing of the information coming from different sources in different languages. Therefore, enhancing tools and resources for under-resourced language is in the core of this initiative activities.

The aim of this article is to present this project focusing on the main strategy behind the development of new resources for under-resourced EU languages in terms of Language Processing Chains (LPC's). The paper is organized as follows: the core CLEOPATRA Knowledge Processing Pipeline (CKPP) will be presented in section 2. In section 3, the role of NLP treatments through LPC's will be described, while the section 4 will encompass a brief analysis of the state of the art of basic tools for under-resourced EU languages. In section 5, the strategy for building enhanced LPC's will be detailed, while the Section 6 will come up with the conclusion and possible future steps

## 2. CLEOPATRA Knowledge Processing Pipeline (CKPP)

The Cleopatra Knowledge Processing Pipeline is composed of four steps as presented in Figure 1 and it comprises the whole event-centric analytics multilingual processing.

The first step refers to the extraction and alignment of event-related information from the varied multilingual media sources to obtain data providing enough linguistic information that will allow further analytics. It concerns the main Language Processing Chains that include tasks from tokenization to parsing and Named Entity Recognition and Classification.

Validation and contextualization of extracted data are part of the second step. For this, textual and visual information will be used to provide fact validation, relation between text and image and sentiment analysis.

---

[1] CLEOPATRA is the acronym for "Cross-lingual Event-centric Open Analytics Research Academy". Website: http://cleopatra-project.eu/

[2] https://ec.europa.eu/eurostat/home?
[3] https://europa.eu/european-union/about-eu/eu-languages_en

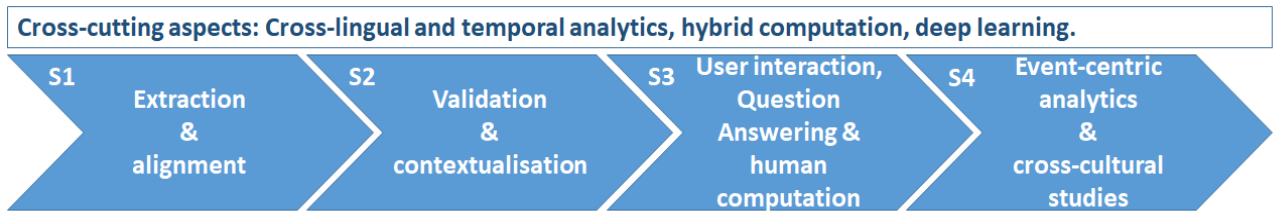

Figure 1: Steps of the Cleopatra Knowledge Processing Pipeline.

The third step involves user interaction with multilingual information to provide an efficient and intuitive search engine relying on the extracted information. Question Answering methods will also be developed to guarantee an effective cross-lingual analysis.

The aim of the fourth and final step of the CKPP is to provide examples of several analytics applications of the pipeline with respect to information propagation and bias, to conduct case studies in politics and sports topical areas, and to analyse community created data sources.

This article will focus on the Language Processing Chains inside steps one and two, which will be responsible for enriching the multilingual data with linguistic information.

### 3. LPCs inside the CKPP

#### 3.1 From Tokenisation to Parsing and Named Entity Recognition and Classification

The main NLP tasks that will be considered are: Sentence splitting, Tokenisation, Lemmatisation, Part-of-Speech and Morphosyntactic tagging, Parsing and NERC. The idea is to provide an online multilingual platform with different tools and pre-trained models that will allow to analyse raw texts from the twenty-four official languages of the European Union.

The platform will contain, for each language, enhanced LPC composed by different existing and newly modules, combining different NLP strategies and methods for different languages aligned with the BLARK (Basic LAnguage Resource Kit)/ELARK (Extended LAnguage Resource Kit) concept defined by ELRA and CLARIN (Krauwer 1998, Maegaard et al. 2005, Arppe et al. 2010). Such an example of NLP online platform is the Web-based Linguistic Chaining Tool (WebLicht) website[4] developed by the German partners in the CLARIN-ERIC, proposing different NLP pre-trained modules that can be combined to annotate texts from forty-one different world-wide languages. Our platform will differ from WebLicht as we aim to propose enhanced processing chains focusing on event-centric media data and to offer new optimized tools using deep learning for under-resourced languages.

For the tasks concerning tokenisation to parsing, in terms of reliable and homogeneous linguistic information to be used as training, development and test data, we will rely mostly on the corpora provided by the Universal Dependencies[5] framework (UD) and use the CoNLL-U format.

All twenty-four official EU languages have available corpora inside UD, however, the amount of data varies enormously between them. This will be analysed further in this article.

For the evaluation of the different LPC's in order to determine the most optimized one, we will follow the standard metrics described in the CoNLL 2018 Shared Task.

The proposed metrics concern each task of the chain individually but also include some combined metrics such as MLAS and BLEX (Straka et al., 2016).

The task of NERC does not have an equivalent framework to the Universal Dependencies one. Instead, many different tools propose different corpora and different types of classification schemes with different complexity in levels and number of predefined categories. However, to guarantee some homogeneity inside the platform to be created, in the beginning, we will base our annotations according to the guidelines of the Seventh Message Understanding Conference (MUC-7) as presented in Mikheev et al. (1997), while the proposal for Universal NER (UNER) scheme is presented in Alves et al. (in press).

#### 3.2 Cross-lingual Sentiment Analysis

For the CKPP, we would like to associate sentiment values with the entities playing pivotal roles in the event. The aim is to process text using the various tools defined in the previous section and perform subjective analysis of the same.

We plan to run this process in two levels of granularities. The first cycle would be sentence-level sentiment analysis and second would be aspect/concept-based sentiment analysis. We assume 3-class (positive, neutral and negative) and 5 class (very-negative, negative, neutral, positive and very positive) classification schemes for sentence-based and aspect-based respectively. The former is easier to begin with and less-prone to phenomena like class-imbalance.

The main Sentiment Analysis tasks that will be considered are: Subjectivity Detection and Subjectivity Classification. For aspect-based sentiment analysis additional detection steps for Opinion Target/s (Aspects/Entities), Opinion Holders, Sentiment Phrase and its classification are foreseen.

Since not all the languages under study have uniform distribution for training data, our main focus in this study would be to employ cross-lingual knowledge transfer methods that have shown good performance on some NLP tasks (Chen et al., 2018; Chidambaram et al., 2019). The creation of resources via the means of crowd-sourcing (implicit as well as explicit) (Nakov et al., 2016) is another viable option in the absence of annotated resources. The systems will be reported in terms of Accuracy, Precision-Recall, and F-Score.

---

[4] https://weblicht.sfs.uni-tuebingen.de/weblichtwiki/index.php/Main_Page

[5] https://universaldependencies.org/

## 4. Under-resourced European Union Languages

As previously mentioned, one of the main challenges when proposing enhanced and robust LPC's in a multilingual platform is the different status of each language in terms of language resources.

According to the META-NET Language Whitepaper series[6] (META-NET LWS 2012), which described a state of the art of NLP development for 31 European languages, there is an enormous discrepancy between them concerning the availability of languages resources and processing tools.

In the following subsections, META-NET Classification will be presented as well as other State of the Art information that allows us to identify the languages that can be considered as under-resourced ones between the EU official languages.

### 4.1 META-NET White Book Series

In the META-NET White Book series (Rehm et al. 2012) a classification of 30 European languages in four different aspects of the development of the respective language technologies was proposed:
- Machine Translation.
- Speech Processing.
- Text Analysis (tools).
- Speech and text resources (data).

For each aspect languages were sorted into five different levels:
- Excellent.
- Good.
- Moderate.
- Fragmentary.
- Weak / No Support.

For CLEOPATRA ITN, the most relevant META-NET categories are "Text Analysis" and "Speech and Text resources". Therefore, considering as under-resourced the languages that are classified as "Fragmentary" or "Weak / No Support" in at least one of these two aspects, the under-resourced EU languages are: Bulgarian, Croatian, Czech, Danish, Estonian, Finnish, Greek, Hungarian, Irish, Latvian, Lithuanian, Maltese, Polish, Portuguese, Romanian, Slovak, Slovenian, Swedish. Eighteen out of twenty-four EU languages.

It is important to mention, however, that this whitepaper series has been published in 2012 and since then some progress has been made in most of these languages. Therefore, an additional contribution of this work will also be an update of META-NET white-papers series information for the selected languages.

Additionally, these two levels actually encompass a wider range of development of language technologies since both, Czech and Maltese appear there although their developments are quite distinct.

### 4.2 Universal Dependencies Corpora

As Universal Dependencies Corpora will be used as a reference for our project, it is also crucial to analyse the languages in terms of the quantity and size of UD corpora. The following table presents a list of all EU languages and the size of the available data in number of tokens.

| Language | Size (Number of tokens) |
|---|---|
| Irish | 40,572 |
| Hungarian | 42,032 |
| Maltese | 44,162 |
| Greek | 61,773 |
| Lithuanian | 75,403 |
| Danish | 100,733 |
| Slovak | 106,043 |
| Bulgarian | 156,149 |
| Slovenian | 170,158 |
| Croatian | 199,409 |
| Swedish | 206,903 |
| Latvian | 220,536 |
| Dutch | 306,764 |
| Finnish | 377,334 |
| Estonian | 465,055 |
| Polish | 496,682 |
| Portuguese | 530,327 |
| Romanian | 551,932 |
| English | 620,511 |
| Italian | 759,457 |
| Spanish | 993,848 |
| French | 1,124,269 |
| Czech | 2,217,119 |
| German | 3,748,466 |

Table 1: EU Languages and the size of their available UD Corpora (version 2.5)[7].

It is possible to notice that while some languages such as German, Czech and French have more than one million tokens corpora, some under-resourced ones have less than fifty thousand tokens datasets. It is the case for Maltese, Hungarian and Irish.

Considering as under-resourced languages the ones with less than five hundred thousand tokens corpora, sixteen can be classified as such: Irish, Hungarian, Maltese, Greek, Lithuanian, Danish, Slovak, Bulgarian, Slovenian, Croatian, Swedish, Latvian, Dutch, Estonian, Finnish, Polish.

---

[6] http://www.meta-net.eu/whitepapers/overview

[7] https://universaldependencies.org/

In comparison with META-NET information, Portuguese and Romanian are classified differently as they have considerable UD corpora.

An important point to consider when using UD datasets is that while the framework proposes stable and homogeneous guidelines, still, it is possible to identify some heterogeneity comparing different UD corpora of the same language: different number of tags used especially for morphological features but also for part-of-speech and dependency relations and different tokenisation strategies for treating contracted words.

### 4.3 Mainstream NLP tools

For this article, a tool is considered mainstream if it proposes pre-trained models for numerous EU languages concerning multiple NLP tasks (mainly from raw text to dependency parsing).

The selected tools that were tested are: Stanford NLP (Manning et al. 2014), UDPipe (Straka et al. 2016), NLP-Cube (Boros et al. 2018), Freeling (Padró & Staniloysky 2012), OpenNLP and spaCy[8].

Only UDPipe have pre-trained models for all EU languages. StandfordNLP and NLP-Cube do not propose downloadable models for Lithuanian or Maltese. The other tools are more limited in terms of multilingual coverage.

Considering the listed tools and their published results and concerning the evaluation metrics of their available models, it is possible to observe that the official results tend to be quite favorable in most of the cases for tasks before parsing. UAS and LAS metrics show more disparity and, thus, will be used here as a criterion for identifying under-resourced languages. If we consider as well-resourced languages the ones with at least one case where UAS is upper than 90, then, we have the following list of nine under-resourced EU languages: Danish, Estonian, Hungarian, Irish, Latvian, Lithuanian, Maltese, Slovak and Swedish.

Comparing UAS values with UD data size information, it is possible to observe that in almost all cases of language with corpora size below one hundred and fifty thousand tokens, UAS values are lower than 90. The only exception being Greek (low size dataset but UAS higher than 90) and Estonian (higher corpora size but low UAS).

All these criteria used to identify EU under-resourced languages were relevant to define the campaign strategy that will be used to build the LPC's and which will be presented in section 5.

### 4.4 Sentiment Analysis

Unlike the other tasks defined in the LPC's, there exist no open-source tools that handle multilingual sentiment analysis for under-resourced languages. The very essential resources required for performing sentiment analysis are sentiment lexicons, which are composed of words and/or multi-word expressions tagged with sentiment scores. Sentiment lexicon alone cannot achieve state of the art as the presence or the absence of intensifier, negations and sarcasm phenomenon can completely modify the expressed sentiment. Hence this type of resource is necessary but not sufficient. However, there have been various attempts in using Machine Translation (MT) as the core tool in generating or aiding the sentiment analysis process. Nevertheless, these MT systems are also prone to inducing translation errors and semantic shift in the translated text. The data (in-domain and out-domain) used for cross-lingual knowledge transfer play a major role in the final performance due to the inherent divergence present (Demirtas et al. 2013). Hence it would be interesting to study the cross-domain, cross-lingual setup for solving the task of sentiment analysis using the LPCs.

## 5. Campaign Strategy

The strategy that will be adopted can be divided in three parts that will be described in the following subsection. First, the idea is to start by developing LPCs for well-resourced languages, secondly, add gradually new processing chains for languages with lesser available data, and, finally, work on the development of modules for most under-resourced languages and integrate them in the platform.

### 5.1 NLP for Well-resourced Languages

In the first phase of the development of the online LPC platform, the focus will be on languages considered well-resourced ones. Taking into consideration the information presented in the previous section, these languages rich in resources and tools are: Dutch, English, French, German, Italian, Spanish.

Considering the existing tools and datasets, the objective is to analyse how all available resources, with different algorithms and methods, can be combined to optimise the processing of event-centric information.

By focusing on existing tools and models trained with sufficient data, our aim is to understand how well different methods work for different tasks and to identify possible synergies between them.

During this phase, the first processing chains will be shared with future CLEOPATRA users so that the possible enhancements concerning formats and interface will be identified.

### 5.2 Deployment to other Languages

In the second step, the idea is to use all the knowledge acquired during the first phase and apply it wisely in the development of LPC's for languages with less resources than the ones listed in the previous section but which are not considered as the most under-resourced ones: Bulgarian, Croatian, Czech, Finnish, Greek, Polish, Portuguese, Romanian and Slovenian.

Existing tools and data will be used and combined in order to achieve the best possible metrics throughout the whole processing chain.

### 5.3 Development of New Modules for Under-resourced Languages

During the last phase, besides testing existing tools, new models based on different deep learning techniques will be developed to all remaining languages, the ones conditionally considered as the most under-resourced ones in the EU: Danish, Estonian, Hungarian, Irish, Latvian, Lithuanian, Maltese, Slovak and Swedish. The results obtained in this step will allow us to compare all techniques and decide which methodology will prevail, not just for particular module design, but for the whole LPCs.

---
[8] https://spacy.io/

All the results obtained will also allow for a deeper understanding of how different deep learning or statistical approaches deal with specific linguistic phenomena of the listed languages.

## 6. Conclusions and Future Directions

The development of robust and effective multilingual Language Processing Chains is crucial for achieving the main objective of CLEOPATRA ITN as text processing is inside the first step of its Knowledge Processing Pipeline. However, due to the difference in available resources of twenty-four official EU languages, an effective strategy must be put in place. Although there is no exact and unique way of classifying a language as under-resourced, we have proposed the division of EU languages into three different clusters, from languages having a good number of resources and tools to languages that could be called the most under-resourced ones in the EU.

This classification is important in our strategy for the development of LPC's as our idea is to start working with very well-resourced languages in the first step, following by the ones with less amount of language resources and finally, in a final phase, take advantage of all the learnings collected in the previous steps together with testing different machine learning methods to create the optimal LPC's for the languages with the lowest number and size of available resources.

Although this work is, primarily, focused on official European Union languages, the main findings could possibly be applied, in further steps, to other under-resources languages in Europe and worldwide.

## 7. Acknowledgements

The work presented in this paper has received funding from the European Union's Horizon 2020 research and innovation program under the Marie Kłodowska-Curie grant agreement no. 812997 and under the name CLEOPATRA (Cross-lingual Event-centric Open Analytics Research Academy).

## 8. Bibliographical References

## 9. Language Resource References